\documentclass[10pt,twocolumn,letterpaper]{article}

\usepackage{cvpr}              

\usepackage{graphicx}
\usepackage{amsmath}
\usepackage{amssymb}
\usepackage{booktabs}
\usepackage{epsfig}
\usepackage{layouts}

\usepackage[pagebackref,breaklinks,colorlinks]{hyperref}

\usepackage[capitalize]{cleveref}
\crefname{section}{Sec.}{Secs.}
\Crefname{section}{Section}{Sections}
\Crefname{table}{Table}{Tables}
\crefname{table}{Tab.}{Tabs.}

\def\cvprPaperID{20} 
\def\confName{CVPR}
\def\confYear{2022}

\begin{document}

\title{The FathomNet2023 Competition Dataset}

\author{Eric C. Orenstein\\
{\tt\small eorenstein@mbari.org}
\and
Kevin Barnard\\
{\tt\small kbarnard@mbari.org}
\and
Lonny Lundsten\\
{\tt\small lonny@mbari.org}
\and
Genevi\`{e}ve Patterson\\
{\tt\small gpatterson@mbari.org}
\and
Benjamin Woodward\\
{\tt\small benjamin.woodward@cvisionai.com}
\and
Kakani Katija\\
{\tt\small kakani@mbari.org}
}
\maketitle

\begin{abstract}
Ocean scientists have been collecting visual data to study marine organisms for decades. These images and videos are extremely valuable both for basic science and environmental monitoring tasks. There are tools for automatically processing these data, but none that are capable of handling the extreme variability in sample populations, image quality, and habitat characteristics that are common in visual sampling of the ocean. Such distribution shifts can occur over very short physical distances and in narrow time windows. Creating models that are able to recognize when an image or video sequence contains a new organism, an unusual collection of animals, or is otherwise out-of-sample is critical to fully leverage visual data in the ocean. The FathomNet2023 competition dataset presents a realistic scenario where the set of animals in the target data differs from the training data. The challenge is both to identify the organisms in a target image and assess whether it is out-of-sample.   
\end{abstract}

\section{Introduction}
\label{sec:intro}
Our ocean is vast, dynamic, and under-explored; a challenging environment that requires new technologies to study, discover, and monitor the organisms that live in the sea. Imaging is increasingly used by ocean ecologists, marine biologists, and public and private sector resource managers to undertake ever more complex sampling campaigns~\cite{durden2016perspectives}. Like many ecological fields, this explosion in imaging-based techniques has resulted in an unwieldy data problem: there is too much of it for any individual or team to analyze alone \cite{beery2018recognition}. Ocean scientists have over the past ten years worked to automate the processing of underwater image data~\cite{irisson2021machine}.

\begin{figure}[t]
  \centering

  \includegraphics[width=0.8\linewidth]{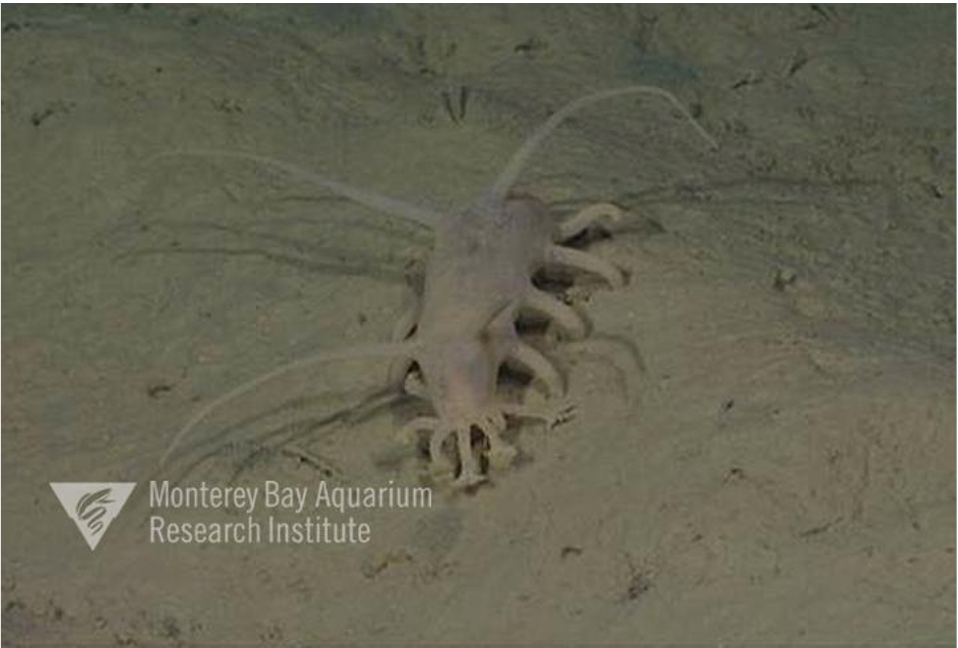}

   \caption{\textit{Scotoplanes} sp. is an organism commonly found below 800 m that is not present in the training dataset. It is very morphologically distinct from concepts above the depth cutoff.}
   \label{fig:seapig}
\end{figure}

While many groups have had success experimenting with machine learning systems to process visual data collected in the ocean, deployments in the field are rare. But these supervised models typically do not generalize to new situations due to distribution shift, when the target distribution some how differs from the training data, a common issue in pattern recognition~\cite{yao2022wild}. This is an especially vexing problem in a dynamic ocean where such shifts can be quite radical: different cameras or illumination used for sampling; novel organisms appearing or known ones disappearing; changes in the appearance of the seafloor, to name just a few~\cite{orenstein2020semi}. Improving robustness and providing tools to identify such changes will allow marine ecologists to better leverage existing data and provide engineers with the ability to deploy instruments in ever more remote parts of the ocean. The FathomNet2023 competition dataset encourages experimentation in this vein by presenting a challenging use case: encountering new organisms and a different species distribution in an unexplored region of the ocean.

\begin{figure}[t]
  \centering

  \includegraphics[width=0.9\linewidth]{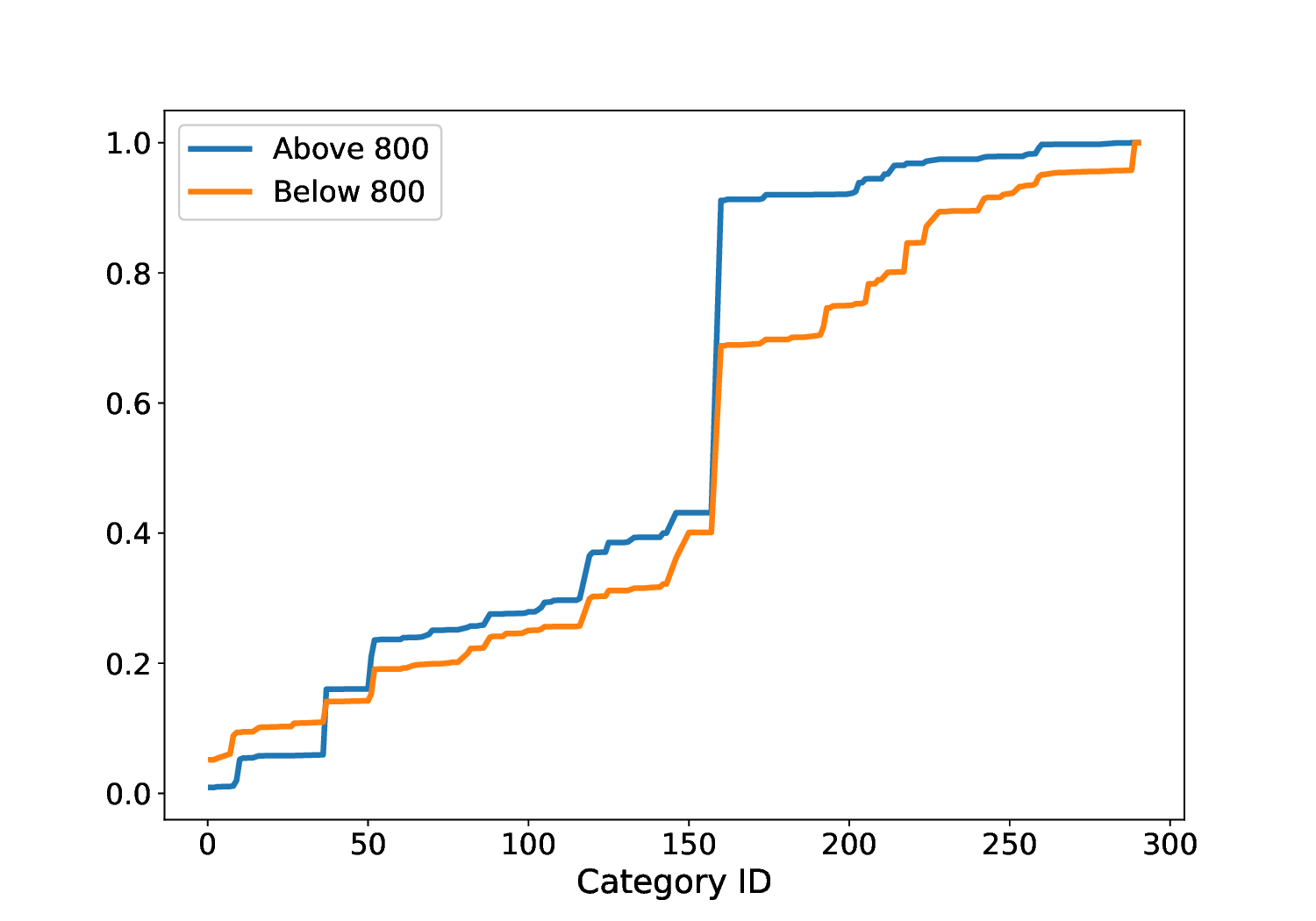}

   \caption{Cumulative distribution of categories in the training and evaluation datasets for FathomNet2023. There are significant differences between them, including some classes that are only present in one set.}
   \label{fig:cumsum-cats}
\end{figure}

For this competition, we have selected data from the broader FathomNet~\cite{katija2022fathomnet} annotated image set, collecting a training set of images collected in the upper ocean (< 800 m) with evaluation data coming from deeper waters. This is a common scenario in ocean research since deeper waters are more difficult to access and, as a result, more annotated data is available closer to the surface. The species distributions in these two regions are overlapping, but not identical, and diverge as the vertical distance between the samples increases (Figs.~\ref{fig:seapig},~\ref{fig:cumsum-cats}). The challenge is both to identify animals in a target image and assess if the image is from a different distribution relative to the training data. 


\section{Dataset Description}
FathomNet is a large-scale repository and database for exploring, sorting, and working with annotated of underwater images~\cite{katija2022fathomnet}. The data it contains are large and diverse, with many potential challenging machine learning problems. For the FathomNet2023 competition dataset, we selected one challenging distribution shift based on depth: training data are drawn exclusively from the upper ocean while the target data contains images from deeper down. 

\subsection{Preparation}
We leveraged FathomNet's rich metadata to isolate the depth shift by restricting the search: (i) geographically to reduce the impact of longitudinal species changes; (ii) institutionally to ensure, to the extent possible, consistency in equipment and personnel; and (iii) temporally to ensure the cameras used for data collection are similar. We also focused our search to bottom-dwelling creatures based on a list of concepts created by regional taxonomic experts. The resulting set of 290 categories represents all benthic fauna captured by the relevant camera systems in the target region.

All images were collected by cameras deployed by the Monterey Bay Aquarium Research Institute (MBARI) on two Remotely Operated Vehicles (ROVs). Data was collected exclusively in the Greater Monterey Bay Area (35.38N to 37.199N, -122.8479W to -121.0046W) between the surface and 1300 meters depth. The images were annotated and localized by domain experts in the MBARI Video Lab using several in-house annotation tools. The training and evaluation data are split across an 800 meter depth threshold: all training data is collected from 0-800 meters, while evaluation data drawn from the whole 0-1300 meter range (Fig.~\ref{fig:cumsum-cats}). The training dataset is composed of images publicly available on FathomNet. The evaluation images were staged for release on FathomNet but were not made public until after the competition.

Annotation data was provided in both multi-label classification and object detection formats. Multi-label classification annotations were presented as a comma separated list image ids and the associated set of unique concepts found in each image. Object detection annotations were formatted to adhere to the COCO Object Detection standard~\cite{lin2014microsoft}, with each image containing at least one localization. Each localization is also annotated as a supercategory in addition the fine-grained labels. The list twenty semantic supercategories group the fine-grained categories morphologically as described by MBARI Video Lab experts. While the fine-grained categories are not all represented in both the training and evaluation data, every supercategory is present in both regions.    

\subsection{Characteristics}
\begin{figure}[t]
  \centering

  \includegraphics[width=0.9\linewidth]{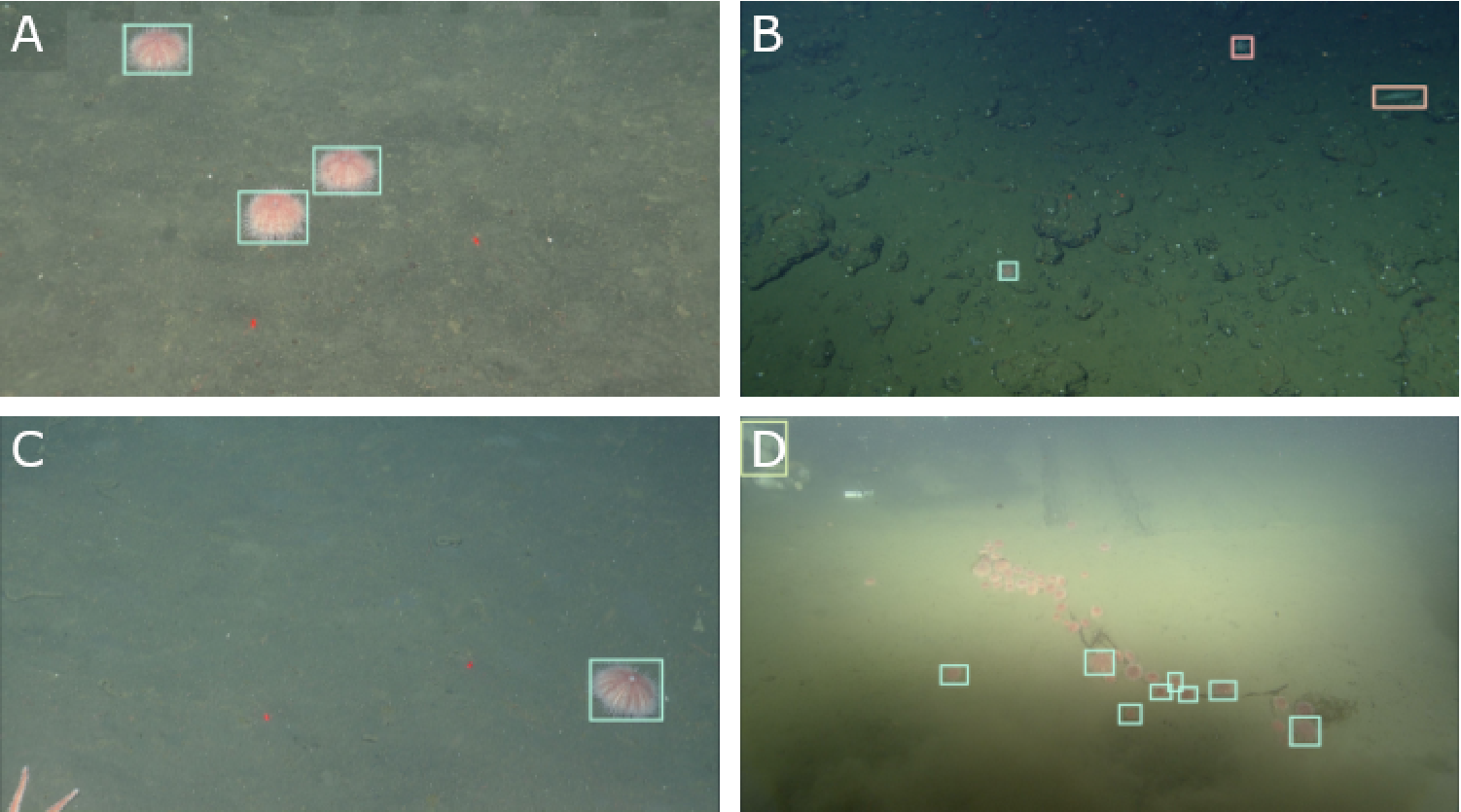}

   \caption{Images of \textit{S. fragilis}, the most common concept in the dataset, that exhibit inherent challenges with deep sea image data. Teal bounding boxes indicate \textit{S. fragilis} in all panels. (A) A completely annotated image with relatively large, in-focus targets. (B) The target is very small relative to the full frame image. (C) The image is blurry due to motion of the vehicle. (D) Not every instance of \textit{S. fragilis} is localized in the frame.}
   \label{fig:issues}
\end{figure}

The FathomNet2023 training dataset is composed of 5950 images containing 23703 localized annotations, while the evaluation data is made up of 10744 images with 49798 localized annotations. 6313 of the evaluation images were collected below the 800 meter threshold and counted as out-of-sample. Like most fine-grained image datasets, the FathomNet2023 dataset is very long-tailed, with a type of urchin (\textit{S. fragilis}) being the most commonly occurring concept in both the training and evaluation set (Figs.~\ref{fig:cumsum-cats},~\ref{fig:issues}A). Beyond \textit{S. fragilis} the order and magnitude of other concepts is quite variable between the sets. 

The images in the competition dataset, and FathomNet as a whole, are subject to several confounding factors that will impact model development and performance. The images represent the natural variability in pose of the animals, containing both iconic and non-iconic views. The target might be occluded, blurry, or small relative to the whole frame (Fig.~\ref{fig:issues}B, C). Some images might not be fully annotated, with individual objects not localized and labeled, resulting in label noise that might challenge object detectors (Fig.~\ref{fig:issues}D).

\subsection{Outside data sources}
We did not provide outside or multi-modal sources but encouraged creative use of other image sets, pre-trained models, taxonomic resources, and species descriptions. Competitors were only restricted from using models shared on the FathomNet model zoo that might contain concepts in the evaluation set. Participants were required to specify any outside resources they consulted when using submitting results. 

\section{Evaluation}
For each image in the test set, competitors are asked to predict a ordered list of categories that appear in the image and classify whether the image is out-of-sample. Results are evaluated on these tasks with mean Average Precision ($mAP$) and Area Under the receiver operating characteristic Curve ($AUC$) respectively. The results are combined for the purposes of ranking on the Kaggle leaderboard. 

\subsection{Out-of-sample}
The receiver operating characteristic (ROC) curve is a visualization of classifier performance at all classification thresholds. The curve plots the true positive and false positive rates at different thresholds where

$$TPR = \frac{TP}{TP+FN}$$

and

$$FPR = \frac{FP}{FP+TN}$$

As the classification threshold gets more permissive, both $FPR$ and $TPR$ increase toward 1. The Area Under the ROC Curve ($AUC$) is computed by integrating the area under the ROC curve and scales between 0 and 1 with a perfect classifier returning $AUC = 1$.  

The out-of-sample predictions are evaluated by with $AUC$. To combine the score with the category evaluation, $AUC$ is rescaled as:

$$sAUC = 2 AUC - 1$$

\subsection{Category predictions}
\label{sec:cat_preds}
Predictions of categories are evaluated according to the mean Average Precision at 20 ($mAP@20$):

$$MAP@20 = \frac{1}{U} \sum_{u=1}^{U} \sum_{k=1}^{min(n,20)} P(k) \times rel(k)$$

where $U$ is the number of images, $P(k)$ is the precision at cutoff $k$, $n$ is the number predictions per image, and $rel(k)$ is an indicator function equaling 1 if the item at rank $k$ is a relevant (correct) label and zero otherwise.

Once a correct label has been scored for an observation, that label is no longer considered relevant for that observation, and additional predictions of that label are skipped in the calculation. For example, if the correct label is `1` for an observation, the following predictions all score an average precision of `1.0`.

\[
 \begin{matrix}
  [1, 2, 3, 4, 5] \\
  [1, 1, 1, 1, 1] \\
  [1, 2, 1, 3, 1]
 \end{matrix}
\]

\subsection{Final score}
The final score is the simple average of $sAUC$ and $mAP@20$:

$$\frac{1}{2} \left( sAUC + MAP@20 \right)$$ 
 

\section{Baseline Results}
\label{sec:baseline}

We developed a simple baseline approach by fine tuning a YOLOv8 object detector on the training dataset toward the 290 fine-grained labels~\cite{yolov8_ultralytics}. The \textit{yolov8m} checkpoint was fine tuned for 50 epochs with an input image size of 640. The resulting model yielded $mAP50 = 0.33$ on a randomly sampled, in-distribution validation set. We stress that $mAP50$ is an object detection specific score that is different from the evaluation metric outlined in Sec. \ref{sec:cat_preds}

The fine-tuned YOLOv8 model was used to detect objects in the evaluation set. We used a common baseline for out-of-sample detection~\cite{hendrycks17baseline}, estimating the metric as the mean confidence over all $N$ detections in an evaluation image:

$$osd = 1 - \frac{1}{N} \sum_{n=1}^{N} {conf}_{n}$$

\begin{figure}[t]
  \centering

  \includegraphics[width=\linewidth]{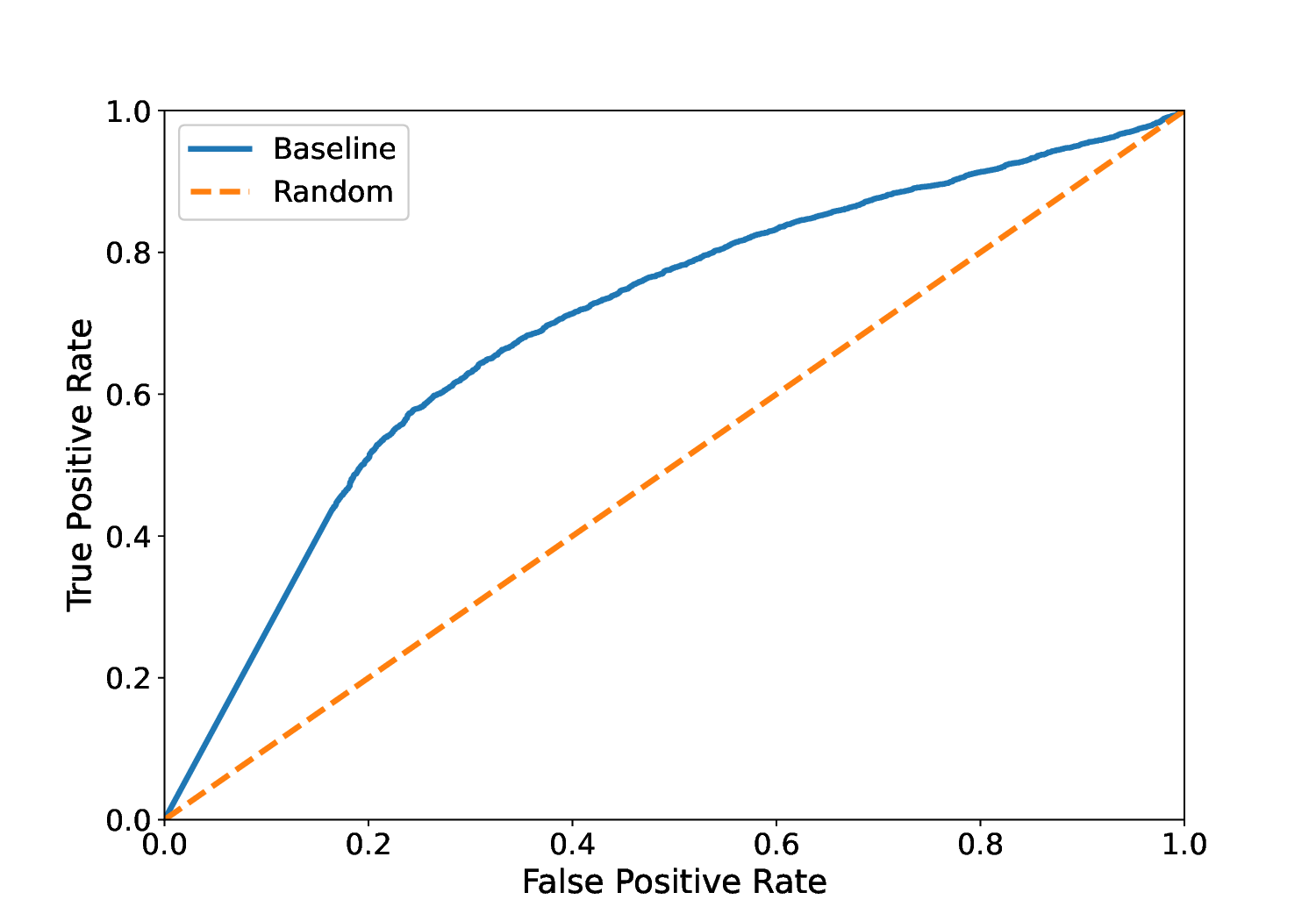}

   \caption{The receiver operating characteristic (ROC) curve of the baseline out-of-sample estimator based on YOLOv8 confidence output. The solid blue line illustrates the detectors performance. The orange dashed line is a visual aid that represents the output of random guess of out-of-sample.}
   \label{fig:roc}
\end{figure}

To conform to the Kaggle submission format, the set of unique detected categories in each frame was used to compute the $mAP@20$ metric. The baseline model returned a $mAP@20 = 0.64$ and $AUC = 0.69$ (Fig.~\ref{fig:roc}), for a final score of $0.26$. 

\section{Conclusion}

The FathomNet2023 dataset is a challenging example of distribution shift in-the-wild. It encodes a common scenario in marine biological observations: species distributions that change rapidly in both space and time. We leveraged the rich metadata available in the FathomNet database to constrain the driver of the shift to depth. The data is presented in standard formats for experiments with both multilabel classification and object detection models. All classes are assigned to supercategories that might be helpful for variety of approaches. While the challenge was focused on out-of-sample detection, the FathomNet2023 dataset could easily be used, for example, in experiments on novel category discovery~\cite{vaze2022generalized}, hierarchical classification~\cite{chen2023taxonomic}, or continual learning~\cite{lin2022continual}. 

The FathomNet database as a whole can be used to create many relevant distribution shifted data challenges: geographically, temporally, or based on sampling equipment. In future years, we are particularly excited to expand the scope of FathomNet challenges by adding multimodal data in the form of plain language descriptions of taxa, relevant scientific texts on the organisms, and coincident environmental data. Such additional will add new dimensions to tackling open-set recognition, model adaptation in new environments, and habitat identification. All of these approaches hold tremendous downstream potential for facilitating responsible use of ocean resources and accelerating discovery in the deep sea. 

\section{Acknowledgements}
We are grateful for the funding support we have received through NOAA-OER, NGS (\#518018), NSF-OTIC (\#1812535),  NSF Convergence Accelerator (ITE \#2137977 and \#2230776) and the David and Lucile Packard Foundation. We are also grateful for the support of the FGVC organizers in helping us prepare and present the competition.  We thank Kaggle for supporting our competition at every stage of the process and featuring it on their main page. 

{\small
\bibliographystyle{ieee_fullname}
\bibliography{bibliography}

\begin{thebibliography}{10}\itemsep=-1pt

\bibitem{beery2018recognition}
Sara Beery, Grant Van~Horn, and Pietro Perona.
\newblock Recognition in terra incognita.
\newblock In {\em Proceedings of the European conference on computer vision
  (ECCV)}, pages 456--473, 2018.

\bibitem{chen2023taxonomic}
Yuzhao Chen, Zonghuan Li, Zhiyuan Hu, and Nuno Vasconcelos.
\newblock Taxonomic class incremental learning.
\newblock {\em arXiv preprint arXiv:2304.05547}, 2023.

\bibitem{durden2016perspectives}
Jennifer~M Durden, Timm Schoening, Franziska Althaus, Ariell Friedman, Rafael
  Garcia, Adrian~G Glover, Jens Greinert, Nancy~Jacobsen Stout, Daniel~OB
  Jones, and Anne Jordt.
\newblock Perspectives in visual imaging for marine biology and ecology: from
  acquisition to understanding.
\newblock {\em Oceanography and Marine Biology: An Annual Review}, 54:1--72,
  2016.

\bibitem{hendrycks17baseline}
Dan Hendrycks and Kevin Gimpel.
\newblock A baseline for detecting misclassified and out-of-distribution
  examples in neural networks.
\newblock {\em Proceedings of International Conference on Learning
  Representations}, 2017.

\bibitem{irisson2021machine}
Jean-Olivier Irisson, Sakina-Doroth{\'e}e Ayata, Dhugal~J Lindsay, Lee
  Karp-Boss, and Lars Stemmann.
\newblock Machine learning for the study of plankton and marine snow from
  images.
\newblock {\em Annual Review of Marine Science}, 14, 2021.

\bibitem{yolov8_ultralytics}
Glenn Jocher, Ayush Chaurasia, and Jing Qiu.
\newblock Ultralytics yolov8, 2023.

\bibitem{katija2022fathomnet}
Kakani Katija, Eric Orenstein, Brian Schlining, Lonny Lundsten, Kevin Barnard,
  Giovanna Sainz, Oceane Boulais, Megan Cromwell, Erin Butler, Benjamin
  Woodward, et~al.
\newblock Fathomnet: A global image database for enabling artificial
  intelligence in the ocean.
\newblock {\em Scientific reports}, 12(1):15914, 2022.

\bibitem{lin2014microsoft}
Tsung-Yi Lin, Michael Maire, Serge Belongie, James Hays, Pietro Perona, Deva
  Ramanan, Piotr Doll{\'a}r, and C~Lawrence Zitnick.
\newblock Microsoft {COCO}: Common objects in context.
\newblock In {\em European conference on computer vision}, pages 740--755.
  Springer, 2014.

\bibitem{lin2022continual}
Zhiqiu Lin, Deepak Pathak, Yu-Xiong Wang, Deva Ramanan, and Shu Kong.
\newblock Continual learning with evolving class ontologies.
\newblock {\em Advances in Neural Information Processing Systems},
  35:7671--7684, 2022.

\bibitem{orenstein2020semi}
Eric~C Orenstein, Kasia~M Kenitz, Paul~LD Roberts, Peter~JS Franks, Jules~S
  Jaffe, and Andrew~D Barton.
\newblock Semi-and fully supervised quantification techniques to improve
  population estimates from machine classifiers.
\newblock {\em Limnology and Oceanography: Methods}, 2020.

\bibitem{vaze2022generalized}
Sagar Vaze, Kai Han, Andrea Vedaldi, and Andrew Zisserman.
\newblock Generalized category discovery.
\newblock In {\em Proceedings of the IEEE/CVF Conference on Computer Vision and
  Pattern Recognition}, pages 7492--7501, 2022.

\bibitem{yao2022wild}
Huaxiu Yao, Caroline Choi, Bochuan Cao, Yoonho Lee, Pang Wei~W Koh, and Chelsea
  Finn.
\newblock Wild-time: A benchmark of in-the-wild distribution shift over time.
\newblock {\em Advances in Neural Information Processing Systems},
  35:10309--10324, 2022.

\end{thebibliography}
}

\end{document}